\documentclass[conference]{IEEEtran}
\IEEEoverridecommandlockouts

\usepackage[ruled,vlined]{algorithm2e}
\usepackage{subfigure}
\usepackage{graphicx}

\usepackage{xcolor}
\usepackage{cite}
\usepackage{amsmath,amssymb,amsfonts}
\usepackage{textcomp}
\usepackage{xcolor}
\def\BibTeX{{\rm B\kern-.05em{\sc i\kern-.025em b}\kern-.08em
    T\kern-.1667em\lower.7ex\hbox{E}\kern-.125emX}}
\begin{document}

\title{Multi-Ship Tracking by Robust Similarity metric}

\author{\IEEEauthorblockN{Hongyu Zhao, Gongming Wei, Yang Xiao, Xianglei Xing*}
\IEEEauthorblockA{\textit{College of Intelligent Science and Engineering} \\
\textit{Harbin Engineering University}\\
Harbin 150001, China \\
xingxl@hrebu.edu.cn \\
*Xianglei Xing is corresponding author}

}

\maketitle

\begin{abstract}
Multi-ship tracking (MST) as a core technology has been proven to be applied to situational awareness at sea and the development of a navigational system for autonomous ships. Despite impressive tracking outcomes achieved by multi-object tracking (MOT) algorithms for pedestrian and vehicle datasets, these models and techniques exhibit poor performance when applied to ship datasets. Intersection of Union (IoU) is the most popular metric for computing similarity used in object tracking. The low frame rates and severe image shake caused by wave turbulence in ship datasets often result in minimal, or even zero, Intersection of Union (IoU) between the predicted and detected bounding boxes. This issue contributes to frequent identity switches of tracked objects, undermining the tracking performance. In this paper, we address the weaknesses of IoU by incorporating the smallest convex shapes that enclose both the predicted and detected bounding boxes. The calculation of the tracking version of IoU (TIoU) metric considers not only the size of the overlapping area between the detection bounding box and the prediction box, but also the similarity of their shapes. Through the integration of the TIoU into state-of-the-art object tracking frameworks, such as DeepSort and ByteTrack, we consistently achieve improvements in the tracking performance of these frameworks. 
\end{abstract}


\section{Introduction}

\begin{figure}[t]
\centering{\includegraphics[scale = 0.69]{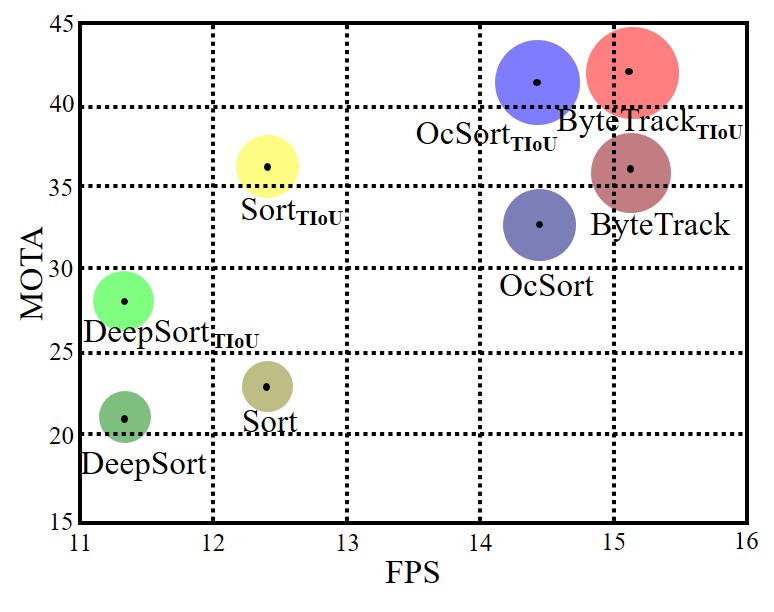}}
\caption{MOTA-IDF1-FPS comparisons of different trackers and their refine versions equipped TIoU. The x-axis represents the frames per second (FPS) which measures the running speed, while the y-axis represents Multiple Object Tracking Accuracy (MOTA). The circle size in the plot is determined by the ratio of detection targets that obtain the correct ID among the targets being detected and tracked (IDF1). The TIoU-equipped trackers have significantly improved their evaluation index in all aspects. Details are given in Table\ref{table}.}
\label{fig}
\end{figure} 

 Multi-object tracking (MOT) technology has been shown to be effective in pedestrian and vehicle datasets. Nonetheless, its success in multi-ship tracking (MST) has been less remarkable. Tracking by detection" is one of the mainstream algorithms in the field of multi-object tracking. To track objects in a video, the Tracking by Detection approach involves two key steps. Firstly, advanced detector algorithms are utilized to obtain a target detection frame for each frame in the video. Secondly, the multi-object tracking process known as data association takes place, whereby tracklets and detection boxes are matched based on their similarity scores, which are computed beforehand. Thus, we focus primarily on improving data association, and \cite{b1} leverage deep learning trends. The Simple Online and Realtime Tracking (SORT) algorithm \cite{b2} employs a straightforward yet effective approach that leverages location and motion cues to track objects in video streams. Specifically, SORT uses a Kalman Filter \cite{b3} to predict the locations of tracklets in the next frame, followed by calculating the Intersection over Union (IoU) similarity between the detection boxes and the predicted locations. Based on the similarity score, objects are then assigned identities using either the Hungarian Algorithm \cite{b4} or a greedy assignment. Another example of object tracking method that improves SORT is ByteTrack \cite{b6}, which utilizes low-score detection boxes to enhance tracking performance. Specifically, it first matches high confidence detections, followed by a second association with low confidence detections.

The Intersection over Union (IoU), which is also referred to as the Jaccard index, is widely used to measure the similarity between two arbitrary shapes. IoU takes into account several properties of the shapes being compared, such as their dimensions and locations, and encodes these properties into a region. By emphasizing the area or volume of these regions, IoU then calculates a normalized measure that provides a reliable estimate of the similarity between the two shapes. In MOT, it is necessary to calculate the IoU between the detection bounding boxes and the prediction bounding boxes as their similarity metric\cite{b7}. However, IoU as a similarity metric in MST has a major issue: due to problems such as poor image quality, low frame rate, and the extremely variable scale between ships, the IoU value can become very small and even zero. As a result, IoU cannot reflect the similarity between the prediction bounding boxes and the detection bounding boxes, which causes serious identity switching. 

In this paper, we aim to overcome the limitation of the IoU by expanding its applicability to scenarios involving minute or non-existent spatial overlap. For predicted bounding boxes $B^{K F}$ and detected bounding boxes $B^{dt}$, we find the smallest enclosing convex object $C$. Compute the ratio of $B^{K F}$ to $C$ and the ratio of $B^{dt}$ to $C$, and then determine the lesser of the two. We introduce this tracking version of IoU, named TIoU, as a new metric for comparing any two arbitrary shapes. TIoU encodes the shape, location, and distance properties of the objects being compared, and then calculates a normalized measure that focuses on their areas. More importantly, TIoU still performs remarkably well in multi-ship tracking, even when the two bounding boxes have little overlap or do not overlap at all. The specific details of the formula and the validity of the theory will be elaborated on in the third section. We have conducted experiments to incorporate TIoU metric into state-of-the-art object tracking algorithms such as DeepSort\cite{b8} and ByteTrack\cite{b6}, the application of the proposed algorithm to the ship dataset resulted in a significant improvement in tracking accuracy.

The paper's main contribution can be summarized as follows:

\begin{itemize}
\item  We introduce this tracking version of IoU, named TIoU, as a new metric for computing the similarity between the detection bounding boxes and the predicted bounding boxes in Multi-ship tracking. 
\item  The calculation of TIoU metric considers not only the size of the overlapping area between the detection bounding box and the prediction box, but also the similarity of their shapes.
\item We incorporate TIoU into the most popular Multi-object tracking algorithms such as DeepSort and ByteTrack and show their performance improvement on Multi-ship tracking. 

\end{itemize}

\section{Related Work}
In this section, we briefly survey relevant works about some improved versions of the IoU as similarity metrics.
\subsection{Intersection over Union}
In this "Tracking-by-Detection" framework, the most important step is to calculate the similarity of the bounding box detected for each frame of the image. The three most commonly used features for data association are motion, location, and appearance features. Some literatures combine two or even three of the three features for similarity calculation.SORT \cite{b22} combines motion cues and location for tracking while DeepSORT \cite{b8} adopts all of three features. Most of the tracking algorithms calculate the Intersection over Union (IoU) between the detection bounding box and the prediction bounding box as the similarity in the motion cues part.

Intersection over Union (IoU) for comparing the similarity between two arbitrary shapes (volumes) $B^{d t}$, $B^{K F}$ $\subseteq \mathbb{S} \in \mathbb{R}^{n}$ is obtained by:
\begin{equation}
I o U=\frac{\left|B^{K F} \cap B^{d t}\right|}{\left|B^{K F} \cup B^{d t}\right|}
\end{equation}
where $B^{d t}=\left(x^{d t}, y^{d t}, w^{d t}, h^{d t}\right)$ is the detection bounding box, and $B=(x^{K F}, y^{K F}, w^{K F}, h^{K F})$ is the tracklet predicted bounding box by Kalman Filter.

\subsection{Generalized Intersection over Union}
Generalized Intersection over Union GIoU\cite{b23} ﬁnds the smallest convex shapes $C \subseteq \mathbb{S} \in \mathbb{R}^{n}$ enclosing both $B^{d t}$ and $B^{K F}$. Then we calculate a ratio between the volume (area) occupied by $C$ excluding $B^{d t}$ and $B^{K F}$ and divide by the total volume (area) occupied by $C$. This represents a normalized measure that focuses on the empty volume (area) between $B^{d t}$ and $B^{K F}$. Finally, GIoU is attained by subtracting this ratio from the IoU value.
\begin{equation}
G I o U =I o U-\frac{\left|C-B^{K F} \cup B^{d t}\right|}{|C|}
\label{eq3}
\end{equation}
where  $C$  is the smallest convex shapes that enclosing both $B^{d t}$ and $B^{K F}$.
\subsection{Distance Intersection over Union}
Distance-Intersection over Union (DIoU)\cite{b24} is another method for measuring the similarity between two bounding boxes. It minimizes the normalized distance between the central points of the boxes.
\begin{equation}
DIoU =I o U-\frac{\rho^{2}\left(\mathbf{b}, \mathbf{b}^{d t}\right)}{c^{2}}
\end{equation}
where $b$ and $b^{d t}$ denote the central points of $B^{K F}$ and $B^{d t}$, $\rho(\cdot)$is the Euclidean distance, and $c$ is the diagonal length of the smallest enclosing box covering the two boxes.

\section{The Proposed Method}

Of course, Intersection over Union (IoU) is the most commonly used similarity calculation metric in multi-object tracking (MOT) benchmarks. The input to calculate the IoU is the predicted location of the tracklets in the new frame by the Kalman filter\cite{b3} and the location of the detection boxes. However, IoU as a similarity metric in MST has a major issue: due to problems such as poor image quality, low frame rate, and the extremely variable scale between ships, the IoU value can become very small and even zero, if $|B^{K F} \cap B^{d t}|=0$, $\operatorname{IoU}(B^{K F}, B^{gt})=0$. In this case, IoU cannot reflect the similarity between the prediction bounding boxes and the detection bounding boxes, which causes serious identity switching.  

And both DIoU and GIoU are used to calculate losses for bounding box regression and non-maximal suppression of redundant detection frames. Their proposed IoU variant is essentially applied to the field of object detection to solve the problem of gradient disappearance in the non-overlapping case and to deal with slow convergence and inaccurate regression. However, it does not address the challenge of data association when the two bounding boxes have only a small overlap or non-overlap in the field of object tracking.

\begin{table}[]
\renewcommand{\arraystretch}{1.6}
\setlength{\tabcolsep}{10pt}
\centering
\caption{Similarity comparison of 4 versions of IoU under 3 tracking situations.}
\label{biao1}
\begin{tabular}{c|c|c|c|c}
\hline
              & IoU                          & GIoU   & DIoU           & {\color{blue}$\mathrm{TIoU}$}                                 \\ \hline
              & 0.753                        & 0.744  & 0.004          & {\color[HTML]{333333} \textbf{0.869}} \\
Large Overlap & {\color[HTML]{333333} 0.730} & 0.719  & 0.005          & {\color[HTML]{333333} \textbf{0.856}} \\
              & 0.681                        & 0.664  & 0.008          & {\color[HTML]{333333} \textbf{0.826}} \\ \hline
              & 0.091                        & 0.176  & 0.131          & {\color[HTML]{333333} \textbf{0.400}} \\
Small Overlap & 0.087                        & 0.194  & 0.141          & {\color[HTML]{333333} \textbf{0.391}} \\
              & 0.034                        & 0.306  & 0.162          & {\color[HTML]{333333} \textbf{0.341}} \\ \hline
              & 0.0                          & -0.500 & \textbf{0.250} & {\color[HTML]{333333} \textbf{0.250}} \\
No Overlap    & 0.0                          & -0.529 & \textbf{0.268} & {\color[HTML]{333333} 0.235}          \\
              & 0.0                          & -0.692 & \textbf{0.356} & {\color[HTML]{333333} 0.154}          \\ \hline
\end{tabular}
\end{table}

In this section, we introduce a new metric called TIoU, which extends IoU to the non-overlapping case. TIoU follows the same definition as IoU by encoding the shape properties of compared objects as area properties, while maintaining the scale-invariant properties of IoU. In addition, TIoU ensures a strong association with IoU for overlapping objects. Using TIoU, we can compare two bounding boxes with little or no overlap, addressing a weakness of IoU.

And then the TIoU function can be deﬁned as:
\begin{equation}
TIoU =\min_{} \left \{ \frac{\left| B^{d t}\right|}{|C|},\frac{\left| B^{K F}\right|}{|C|} \right \}
\end{equation}
where the definition of $C$ is consistent with the definition of $C$ in Equation\ref{eq3}, TIoU as a new similarity metric has the following properties:
\begin{enumerate}
\item	TIoU, like IoU, holds all the properties of a metric, such as non-negativity, the identity of indiscernibles, and symmetry; 
\item	TIoU, like IoU, is scale-invariant;
\item	$\forall B^{d t}, B^{K F} \subseteq \mathbb{S}, 0 \leq \operatorname{IoU}(B^{d t}, B^{K F}) \leq 1$, and TIoU has a symmetric range,i.e.  \ $\forall B^{d t}, B^{K F} \subseteq \mathbb{S},0 \leq\operatorname{TIoU}(B^{d t}, B) \leq 1$:

\item	As with IoU, a score of 1 is achieved only when two bounding boxes are perfectly aligned, i.e. if $|B^{d t} \cup B^{K F}|=|B^{d t} \cap B^{K F}|$, then TIoU = IoU = 1;

\item	The TIoU value converges asymptotically to zero when the ratio of the occupied area of one of the two shapes to the volume (area) of the enclosing shape tends to zero;

\item   TIoU is more sensitive to changes in the shape and aspect ratio (area) of both input detection and predicted bounding boxes, compared to IoU, which is primarily limited to the changes occurring in the intersection region between the two bounding boxes.
\end{enumerate}

\begin{figure}[t]
\centering{\includegraphics[scale = 0.35]{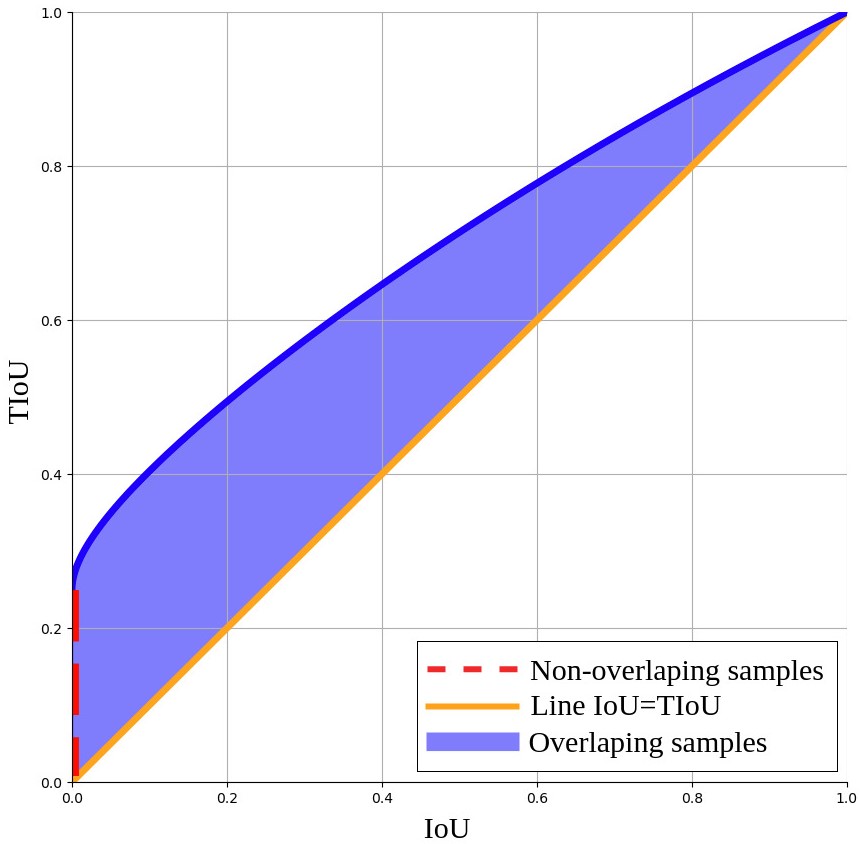}}
\caption{The correlation between TIoU and IoU is measured for overlapped and non-overlapped sample pairs.}
\label{fig2}
\end{figure}
In all non-overlapping cases, IoU has a value of zero, which can lead to inaccurate multi-ship tracking results. In contrast, TIoU metric has a value in all possible cases, including non-overlapping situations. TIoU also shows a strong correlation with IoU, especially in cases where the IoU value is high. This correlation is demonstrated qualitatively in Figure \ref{fig2}, where TIoU exhibits a more significant change than IoU in cases of low overlap. As a result, TIoU has the potential to provide a more comprehensive measure of similarity than IoU. Our experimental results validate this claim.

As shown in Table\ref{biao1}, using the size of the IoU value as a criterion, we classify the possible cases of similarity between the  detection bounding box and  tracklet predicted bounding box by Kalman Filter into three types: large overlap, small overlap, and no overlap, and We select three typical samples of each. Experimental results prove the superiority of TIoU, TIoU is not as good as DIoU only in non-overlapping scenarios, but the calculation logic of DIoU obviously does not conform to the application in MST.

\begin{figure}[htbp]
    \centering
    \includegraphics[scale = 0.55]{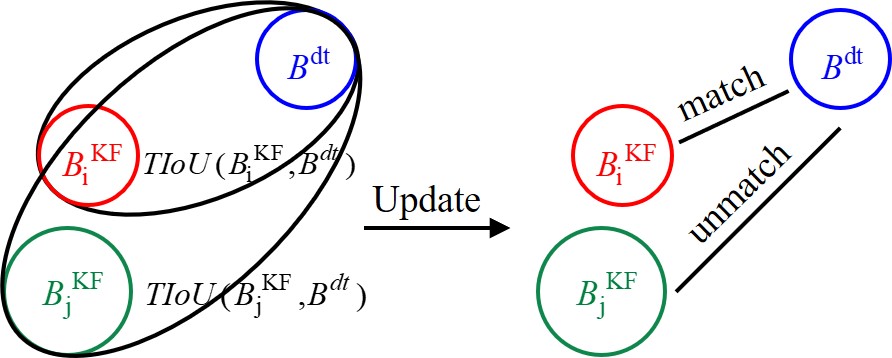}
    \caption{For example, there are two prediction boxes $\mathrm{B}_{\mathrm{i}}^{\mathrm{KF}}$ and $\mathrm{B}_{\mathrm{j}}^{\mathrm{KF}}$, $\mathrm{B}_{\mathrm{i}}^{\mathrm{KF}}$ and detection box $\mathrm{B}^{\mathrm{dt}}$ are not only close to each other but also similar in shape, while $\mathrm{B}_{\mathrm{j}}^{\mathrm{KF}}$ and detection box $\mathrm{B}^{\mathrm{dt}}$ are far away and differ greatly in shape, so $\operatorname{TIoU}\left(\mathrm{B}_{\mathrm{i}}^{\mathrm{KF}}, \mathrm{B}^{\mathrm{dt}}\right)$ $>$ $\operatorname{TIoU}\left(\mathrm{B}_{\mathrm{j}}^{\mathrm{KF}}, \mathrm{B}^{\mathrm{dt}}\right)$, after matching by the Hungarian algorithm, prediction box $\mathrm{B}_{\mathrm{i}}^{\mathrm{KF}}$ and detection box $\mathrm{B}^{\mathrm{dt}}$ are successfully associated.}
    \end{figure}

\begin{figure}[t]
\centering  
\subfigure[$B^{KF} < B^{dt}$]{
\label{a}
\includegraphics[scale = 0.54]{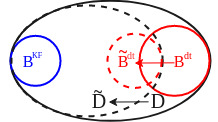}}\subfigure[$B^{dt} < B^{KF}$]{
\label{b}
\includegraphics[scale = 0.54]{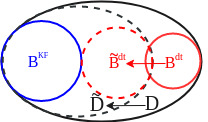}}
\caption{The dynamic process of optimizing TIoU can be visualized.} 
\label{dy}
\end{figure}

From an optimization perspective, we always hope that the TIoU value is as high as possible between the predicted bounding box and the detection bounding box. When $B^{KF}$ is smaller than $B^{dt}$, as equation \ref{e5} suggests, the two most likely characteristics of the detection bounding boxes that match the predicted bounding box are: firstly, they have the same shape and size as the predicted bounding box, and secondly, they have the highest possible overlap with the predicted bounding box. As shown in Fig \ref{dy}:
\begin{equation}
\begin{aligned}\max \left\{{\text { TIoU }}\right\} & =\max \frac{B^{K F}}{C} \\& =\max \frac{B^{K F}}{B^{K F}+B^{d t}+D} \\& =\min \frac{B^{K F}+B^{d t}+D}{B^{K F}} \\& =\min (\frac{B^{d t}}{B^{K F}}+\frac{D}{B^{K F}} + 1) \\& =\min \frac{B^{d t}}{B^{K F}}+\min \frac{D}{B^{K F}} +1\\\end{aligned}
\label{e5}
\end{equation}

where $D$ is the area obtained by subtracting the area of $B^{dt}$ and $B^{KF}$, respectively removed from $C$. When $B^{dt}$ is smaller than $B^{KF}$, as equation \ref{e6} suggests, consistent conclusions will be obtained. The consistency property of TIoU pertains to the fact that it offers a standardized and consistent measure of overlap between two bounding boxes, irrespective of their sizes, shapes, or orientations. 
\begin{equation}
\begin{aligned}\max \left\{\text { TIoU }\right\} & =\max \frac{B^{dt}}{C} \\& =\max \frac{B^{dt}}{B^{K F}+B^{d t}+D} \\& =\min \frac{B^{K F}+B^{d t}+D}{B^{dt}} \\& =\min (\frac{B^{KF}}{B^{dt}}+\frac{D}{B^{dt}} +1) \\& =\min \frac{B^{KF}}{B^{dt}}+\min \frac{D}{B^{dt}} +1\\\end{aligned}
\label{e6}
\end{equation}

In conclusion, TIoU retains the critical characteristics of IoU while addressing its limitations, making it a suitable alternative for all performance measures utilized in 2D/3D computer vision applications. Our current focus has been on Multi-ship tracking, where an analytical solution called TIoU can be easily derived and applied as a metric. We plan to explore suitable similarity metrics in 3D tracking in future work.

\section{Experiments and analysis}
\subsection{Datasets}
Marine datasets are relatively scarce in the research community due to the commercial or military nature of most marine applications. Thanks to the "Jereh Cup" sea surface target detection and tracking dataset created by the Jiangsu Institute of Automation, our group have participated in the annotation of the dataset, and the specific annotation format is the same as MOT17\cite{b25}. All the images were captured at a resolution of 1080 × 1920. There are 35 video sequences used as a training set with 29758 images and 45 video sequences used as a test set with 27812 images. The "Jereh Cup" sea surface target detection and tracking dataset includes four real sea scenarios: port area, outbound port, inbound port, and offshore, among which offshore target tracking is difficult. Considering the types and motion characteristics of the surface targets, the dataset is divided into seven categories, including sailboats, cargo ships, and speedboats. Considering the problems faced in the detection and tracking task, such as unbalanced target categories, mutual target occlusion, small target size, and difficulty in target association under complex lighting or carrier maneuvering conditions, the dataset is designed by selecting typical sea surface scenarios.

\subsection{Metrics}
We use the CLEAR metrics to evaluate different aspects of tracking performance. These metrics include $IDF1$, which evaluates the identity preservation ability and focuses on the association's performance. We also use $Recall$, which measures the percentage of detected objects compared to the ground truth objects, and $MOTP$, which indicates the overlaps between the predicted locations and ground truth locations. False positives are measured using $FP$, while false negatives are measured using $FN$. The identification switch is evaluated using $IDS$, and trajectory FM is evaluated using $FM$. Finally, we use $MOTA$, which is a measure of MOT accuracy that combines IDS, FP, and FN, and focuses more on the detection performance.

\subsection{Implementation details}
For TIoU, as shown in algorithm\ref{a1}, it is only necessary to replace IoU with the definition of TIoU in the tracking algorithm to complete an efficient MST. If TIoU between a detected bounding box and the predicted box of a tracklet is less than 0.1, the matching will be rejected.

\begin{figure*}[t]
\centering  
\subfigure[Offshore sea state scenes]{
\label{demo1}
\includegraphics[scale = 0.14]{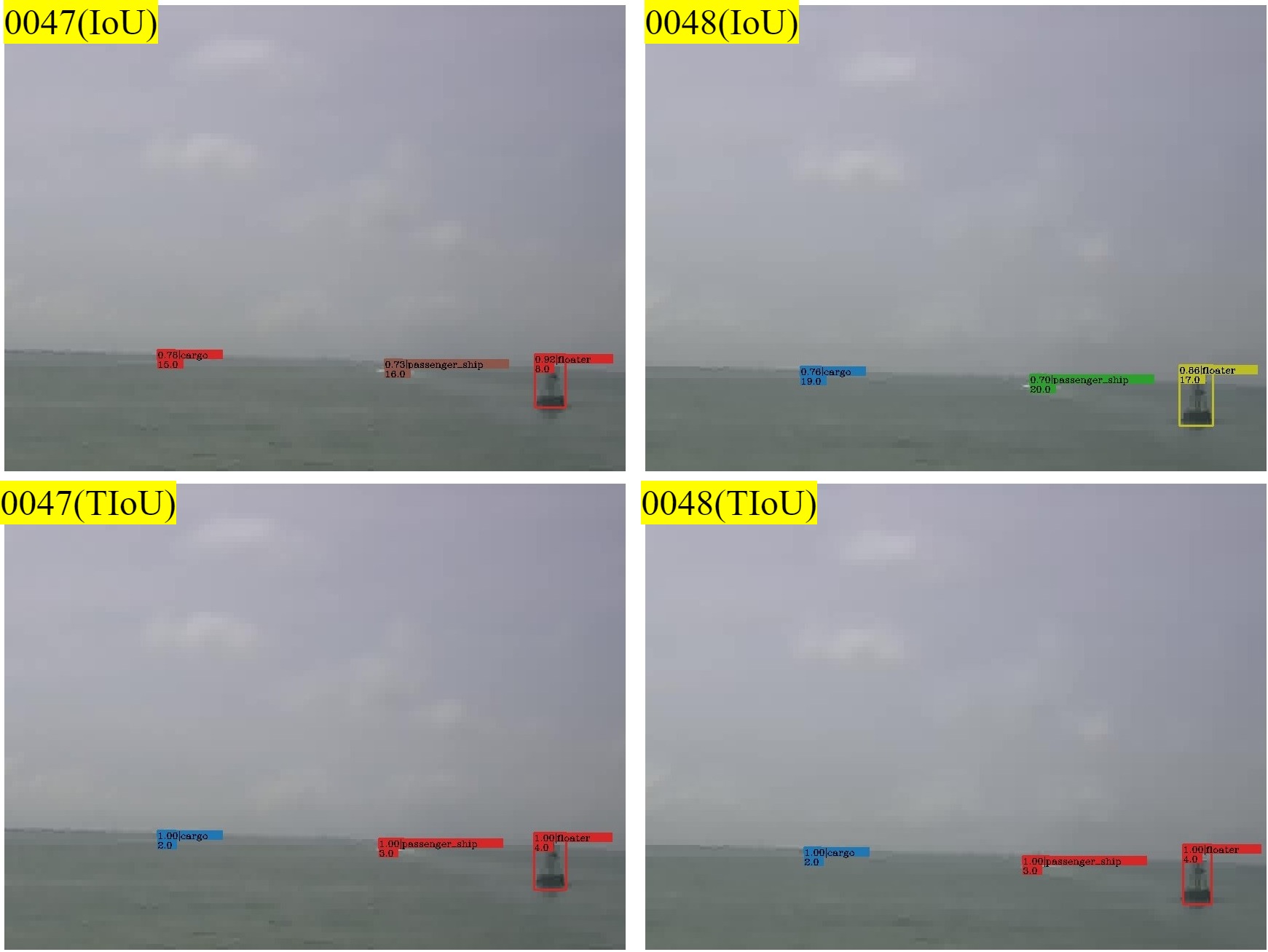}}\subfigure[Sea state scenes in the port area]{
\label{demo2}
\includegraphics[scale = 0.14]{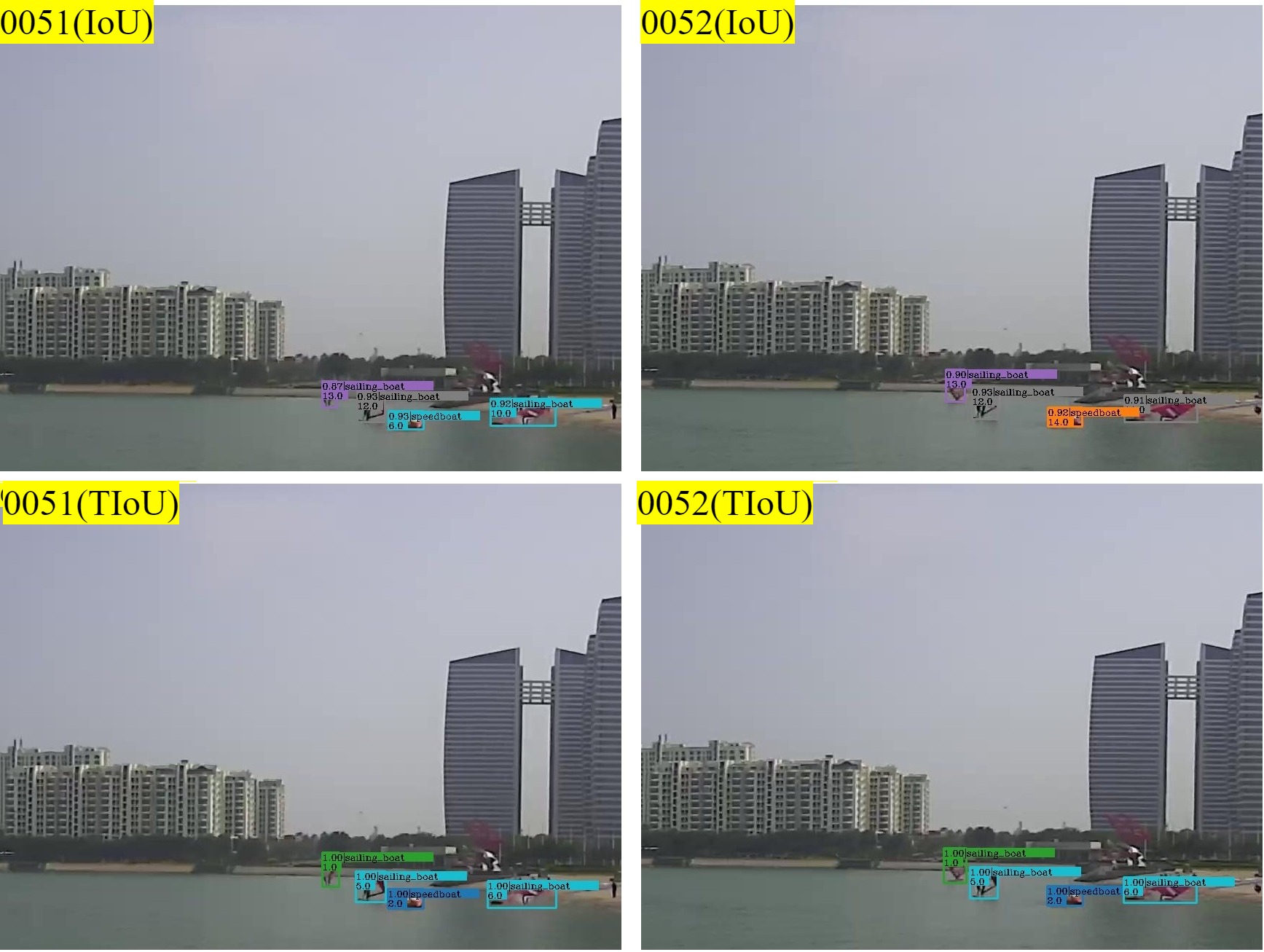}}
\caption{Tracking examples utilizing IoU and TIoU on a different scene, and the number in the upper left corner represents the number of frames, and in parentheses is the metric for calculating similarity.}
\label{demo}
\end{figure*}

\begin{figure*}[htp]
\centering
\includegraphics[scale=0.6]{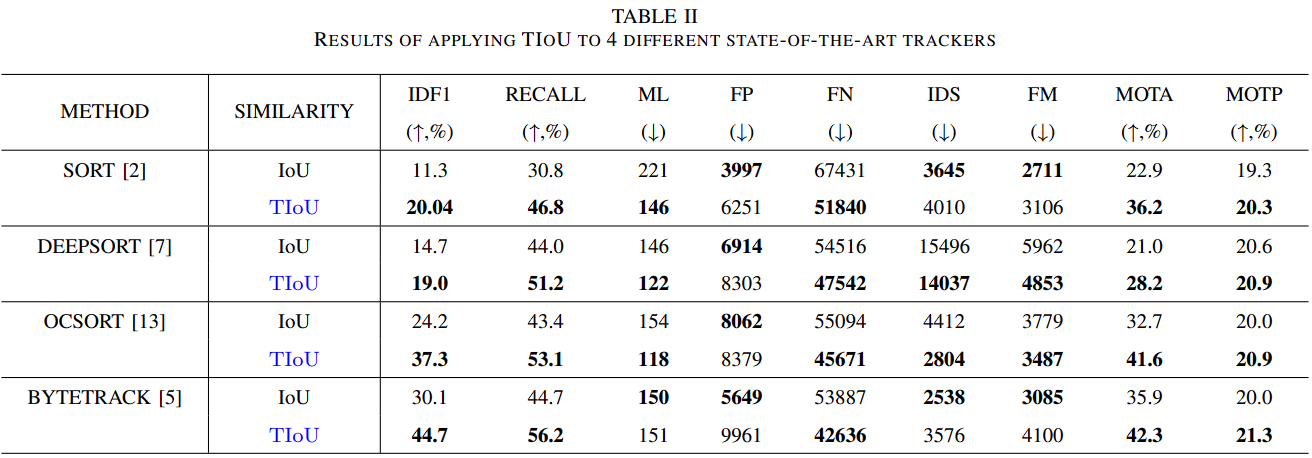}  
\label{table}    
\end{figure*}

The YOLOX\cite{b27} detector with YOLOX-X backbone is trained on the 'Jereh Cup' sea surface target detection and tracking dataset for 120 epochs, with input image size of 1440 x 800. Data augmentation techniques, Mosaic and Mixup, are used, and the model is trained on four NVIDIA 2080ti GPUs with 4 samples per batch, for a total training time of approximately ten days. The optimizer used for training is Stochastic Gradient Descent (SGD) with a weight decay of $5 \times 10^{-4}$ and a momentum of 0.9. The initial learning rate is set to $2.5 \times 10^{-4}$. FPS is measured following the methodology outlined in \cite{b27} with FP16 precision and we test on a single 2080ti with a batch size of 1. The code and results can be found at our github project. We have implemented the algorithm with the MindSpore framework.

\subsection{\textbf{Ablation Studies on TIoU}}
We apply TIoU on 4 different state-of-the-arts trackers, including SORT\cite{b2}, DeepSort\cite{b8}, OcSort\cite{b28} and ByteTrack\cite{b6}. The results are shown in Table\ref{table}.

SORT can be seen as a baseline approach to the "Tracking by detection" tracking strategy. We can see that TIoU improves the MOTA metric of SORT from 22.9 to 36.2, IDF1 from 11.3 to 20.4, and decreases ML from 221 to 146, and FN from 67431 to 51840. This highlights the importance of the TIoU metric to compensate for prediction loss in cases where the low frame rate of the dataset and wave perturbation conditions lead to inaccurate Kalman filter prediction targets.

DeepSORT uses additional Re-ID models to enhance the long-range association. We can see that TIoU improves the MOTA metric of DeepSORT from 21.0 to 28.2, IDF1 from 14.7 to 19.0 and decreases ML from 146 to 122, and FN from 54516 to 47542. The high number of IDSs in tracking tasks presents a significant challenge to association algorithms. However, we have observed that a simple Kalman Filter combined with TIoU can effectively perform long-range association and achieve better IDF1 and MOTA scores, especially when detection boxes are sufficiently accurate. It is worth noting that in situations where ships become heavily obscured, Re-ID features may become less reliable and lead to more identity switches. In contrast, motion models tend to behave more reliably in such scenarios.

OcSort corrects the Kalman filter. The tracking metrics of OcSort equipped with TIoU are also significantly improved. OcSort also introduces the principle of directional consistency in the calculation of IoU, which proves that our improvement of IoU is in the right direction.

ByteTrack sets the data association of high and low-score detection boxes with tracklets respectively. We introduce the concept of TIoU in these two association processes, finally, the final results demonstrate that TIoU can successfully re-associate previously unmatched tracklets. This results in more accurate tracklet boxes.

\begin{figure}[htp]
\centering
\includegraphics[scale=0.6]{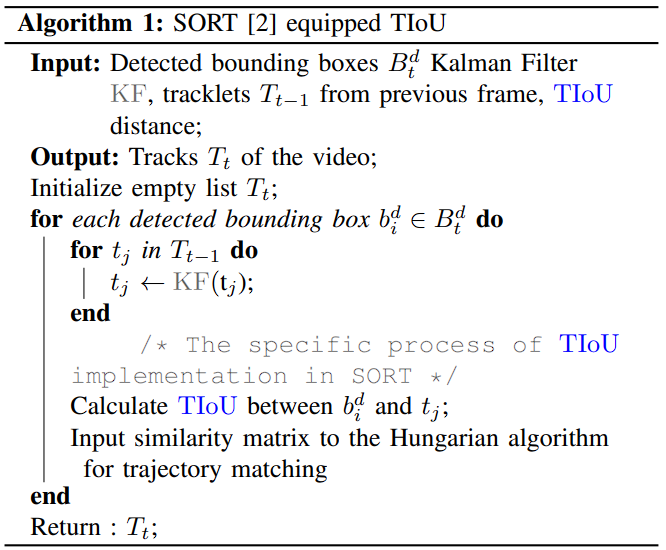}  
\label{a1}    
\end{figure}
Fig\ref{demo} represents four tracking examples utilizing IoU and TIoU as similarity metrics, which ByteTrack is used as baseline method. As shown, due to sea waves and low image frame rates, the tracking results of IoU, the IDs of all the ships in Fig\ref{demo1} at frame 47 are incorrectly assigned at frame 48, however, under TIoU, no switching occurs (all three ships are always regarded as distinct). 
In Fig\ref{demo2}, comparable tracking results are shown. TIoU can accurately locate the positions of ships in subsequent frames.

\subsection{\textbf{Shape Sensitivity of TIoU}}

\begin{figure}[t]
\centering  
\subfigure[$IoU$ = 0.143  $TIoU$ = 0.444]{
\label{a}
\includegraphics[scale = 0.52]{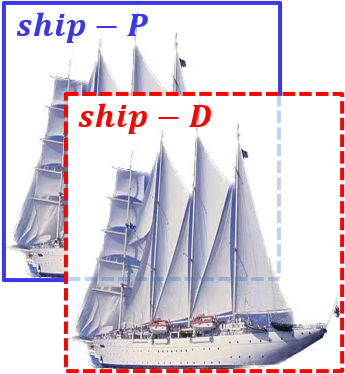}}\subfigure[$IoU$ = 0.143  $TIoU$ = 0.4]{
\label{b}
\includegraphics[scale = 0.52]{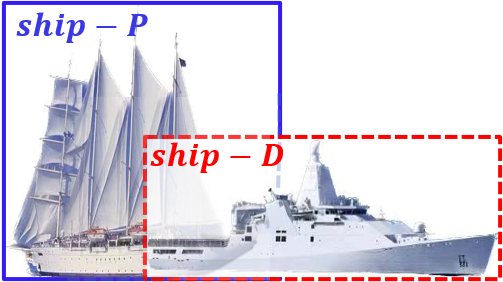}}
\caption{{\color{red}Red} is the detection bounding box, {\color{blue}blue} is the tracklet predicted bounding box. For the same IoU value, the TIoU value of two boxes with similar shapes will be larger, which helps to achieve correct tracking.} 
\label{Fig4}
\end{figure}

IoU only determines the area with the detection boxes and the prediction boxes. However, TIoU has a natural characteristic compared to IoU: it will be sensitive to the shape of the detection boxes and the prediction boxes. Intuitively, the prediction boxes and the detection boxes are likely to match correctly only if they are almost similar in shape. As shown in Fig\ref{Fig4}, Fig\ref{a} and Fig\ref{b} have the same IoU value, which can cause problems for the subsequent Hungarian algorithm and thus lead to IDS. However, if TIoU is used as the similarity metric, the difference in shape can be correctly distinguished.

\section{Conclusion}

In this work, we proposed TIoU, a tracking version of the Intersection over Union metric, to measure the similarity between the predicted boxes and detection boxes. We demonstrated that TIoU retains all of the desirable properties of IoU, while also correcting its limitations. As a result, TIoU can be considered a reliable substitute for IoU in various performance evaluation metrics used in multi-ship tracking applications.

Our proposed data association method, TIoU, offers a straightforward yet powerful solution to the multi-ship tracking problem. The proposed method readily integrates into existing trackers and presents consistent improvements in multiple tracking metrics such as MOTA, IDF1, and Recall. TIoU achieves these results by enabling the Kalman Filter to generate smoother predictions and enhancing the tracker's association ability. The simplicity and speed of TIoU make it an appealing option for practical ship tracking applications. Overall, our study shows the potential of TIoU to enhance tracking performance and improve the accuracy of computer vision systems.

\section*{Acknowledgements}
The work of was supported by the Natural Science Foundation of China No. 61703119 and the CAAI-Huawei MindSpore Open Fund No. CAAIXSJLJJ-2020-033A.

\end{document}